\documentclass[runningheads]{llncs}
\usepackage{graphicx}
\usepackage{amsmath,amssymb} % define this before the line numbering.
\usepackage{color}

\usepackage{bm}
\usepackage{subcaption}
\usepackage[colorlinks]{hyperref}
\newcommand{\eg}{\textit{e.g.}}
\newcommand{\ie}{\textit{i.e.}}

\usepackage{hyperref}
\begin{document}
% \linenumbers
\pagestyle{headings}
\mainmatter
\def\ECCV18SubNumber{1372} 

\title{Shuffle-Then-Assemble: Learning Object-Agnostic Visual Relationship Features} 
% Replace with your title

\titlerunning{Shuffle-Then-Assemble}
% Replace with a meaningful short version of your title

\authorrunning{Xu Yang, Hanwang Zhang, Jianfei Cai}
% Replace with shorter version of the author list. If there are more authors than fits a line, please use A. Author et al.

\author{Xu Yang, Hanwang Zhang, Jianfei Cai}

%Please write out author names in full in the paper, i.e. full given and family names. 
%If any authors have names that can be parsed into FirstName LastName in multiple ways, please include the correct parsing, in a comment to the volume editors:
%\index{Lastnames, Firstnames}
%(Do not uncomment it, because you may introduce extra index items if you do that, we will use scripts for introducing index entries...)

\institute{ School of Computer Science and Engineering,\\
	Nanyang Technological University, \\
	\email{ s170018@e.ntu.edu.sg,\{hanwangzhang,asjfcai\}@ntu.edu.sg}
}

\maketitle

\begin{abstract}
Due to the fact that it is prohibitively expensive to completely annotate visual relationships, \ie, the (obj1, rel, obj2) triplets, relationship models are inevitably biased to object classes of limited pairwise patterns, leading to poor generalization to rare or unseen object combinations. Therefore, we are interested in learning object-agnostic visual features for more generalizable relationship models. By ``agnostic'', we mean that the feature is less likely biased to the classes of paired objects. To alleviate the bias, we propose a novel \texttt{Shuffle-Then-Assemble} pre-training strategy. First, we discard all the triplet relationship annotations in an image, leaving two unpaired object domains without obj1-obj2 alignment. Then, our feature learning is to recover possible obj1-obj2 pairs. In particular, we design a cycle of residual transformations between the two domains, to capture shared but not object-specific visual patterns. Extensive experiments on two visual relationship benchmarks show that by using our pre-trained features, naive relationship models can be consistently improved and even outperform other state-of-the-art relationship models. 
Code has been made available at: \url{https://github.com/yangxuntu/vrd}.
\end{abstract}

\section{Introduction}
Thanks to the maturity of mid-level vision solutions such as object classification and detection~\cite{he2016deep,redmon2016yolo9000,gu2015recent}, we are more ambitious to pursue higher-level vision-language tasks such as image captioning~\cite{gu2017stack,gu2017empirical,chen2017sca,liu2018contextaware}, visual Q\&A~\cite{jabri2016revisiting,li2018vqa,gurari2018vizwiz}, and visual chatbot~\cite{das2017visual}. Unfortunately, we gradually realize that many of the state-of-the-art systems merely capture training set bias while not the underlying reasoning~\cite{vinyals2017show,jabri2016revisiting,zhou2015simple}. Recently, a promising way is to use visual compositions such as scene graph~\cite{johnson2015image,xu2017scene} and relationship context~\cite{hu2016modeling,zhang2018grounding} for explainable visual reasoning. Therefore, visual relationship detection (VRD)~\cite{zhang2017visual,zhang2017ppr,li2017vip,yin2018zoom} --- the task of predicting elementary triplets such as ``person ride bike'' and ``car park on road'' in an image --- is becoming an indispensable building block connecting vision with language.

\begin{figure}[t!]
\centering
\includegraphics[height=3.8cm]{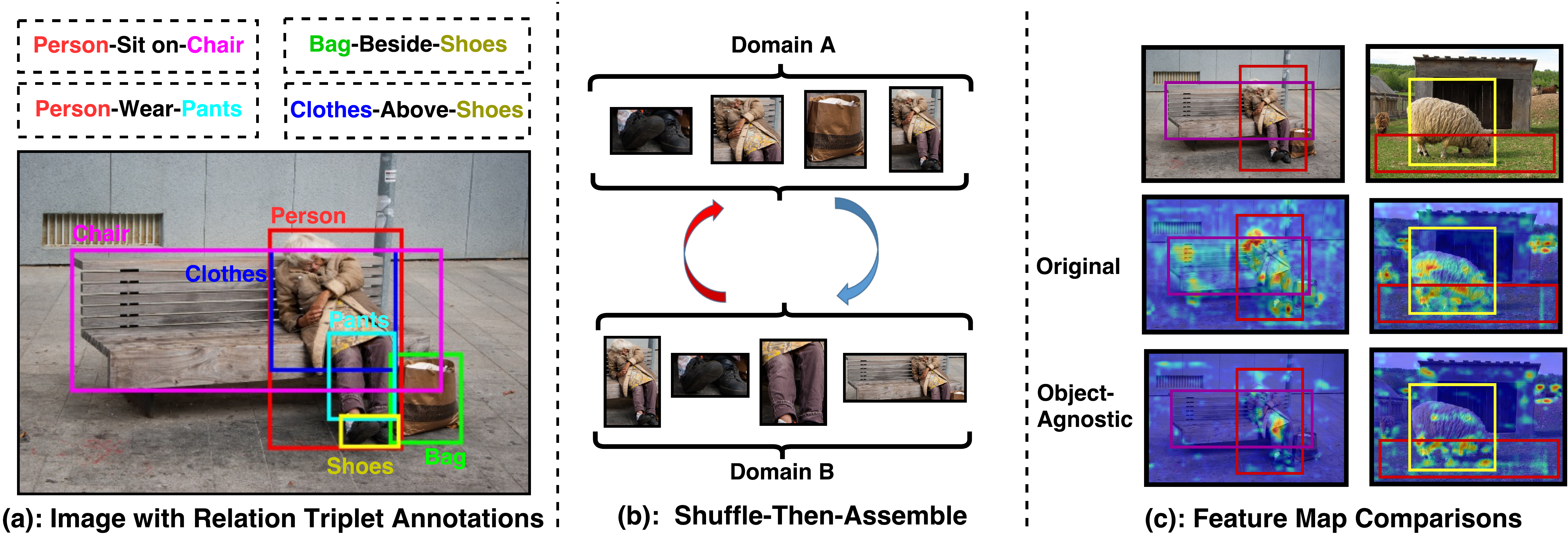}
\caption{(a) The triplet annotations of visual relationships in an image. (b) The key idea of the proposed \texttt{Shuffle-Then-Assemble} strategy is to discard the paired annotation of any relationship and leave two unpaired object domains. (c) Comparisons between the original feature maps obtained from base CNN (middle) and the object-agnostic ones (bottom) obtained by our pre-training (averaged over all the channels). We can see that our feature maps focus more on the overlapped regions of objects.}
\label{fig:1}
\end{figure}

Despite the relatively preliminary stage of VRD compared to object detection, a major challenge of VRD is the high cost of annotating the (obj1, rel, obj2) triplets as shown in Fig.~\ref{fig:1} (a). Unlike labeling objects in images, labeling visual relationships is prohibitively expensive as it requires combinatorial checks of the three entries. Therefore, the relationships in existing VRD datasets~\cite{lu2016visual,krishna2017visual} are long-tailed, and the resultant relationship models are inevitably biased to the dominant obj1-obj2 combinations. For example, as reported in pioneering works~\cite{zhang2017visual,lu2016visual}, the recognition rate of unseen triplet compositions is significantly lower than the seen ones. This deficiency clearly limits the VRD potential in compositional reasoning. Though it can be alleviated by exploiting external knowledge such as language priors~\cite{lu2016visual} and large-scale weak supervision~\cite{zhang2017ppr}, we still lack a principled solution in the visual modeling perspective.

Unsupervised feature learning (or pre-training) is arguably the most popular remedy for training deep models with small data~\cite{zhang2017split,noroozi2016unsupervised,doersch2015unsupervised,radford2015unsupervised,donahue2016adversarial,tao2017zero}. Therefore, we are inspired to learn object-agnostic convolutional feature maps that are less likely biased to certain obj1-obj2 combinations. Such features should be highly responsive to object parts\footnote[1]{The parts can be at the pixel-level as well as the receptive field-level.} involved in a relationship.  A plausible way is to append additional conv-layers to the original base CNN (\eg, VGG16~\cite{simonyan2014very} or ResNet-150~\cite{he2016deep}) to remove the object-sensitive responses inherited from image classification pre-training dataset (\eg, ImageNet~\cite{deng2009imagenet}). For example, as shown in Fig.~\ref{fig:1} (c), compared with the base CNN's feature map, the object-agnostic one ignores object patterns but focuses on the shared patterns of interacted objects. Therefore, we raise a question: how to learn the object-agnostic feature maps without additional relationship labeling cost?

In this paper, we propose a novel \texttt{Shuffle-Then-Assemble} feature learning strategy. As shown in Fig.~\ref{fig:1} (b), ``shuffle'' is to discard the original one-to-one paired object alignments, and thus no explicit obj1-obj2 class information is used; ``assemble'' is to pose the relationship modeling into an unsupervised pair recover problem by transferring Region-of-Interest (ROI) features between the two unpaired domains. Our intuitive motivation is two-fold: 1) if the ROI features extracted from the resultant feature maps still encode object-specific information, features are not likely to be transferred between the two domains of heterogeneous objects; 2) the unsupervised fashion encourages the exploration of many more possible relationships which are usually missing in annotation. As shown in Fig.~\ref{fig:1} (a), some simple spatial relationships such as ``chair beside bag'' are missing, and equivalent relationships are usually ignored, \ie, ``chair under person'' is missing as ``person sit on chair'' is labeled. Inspired by the recent advances in unsupervised domain transfer methods~\cite{zhu2017unpaired,kim2017learning,hoffman2017cycada,yi2017dualgan}, we design a cycle of transformations to establish the transfer between the two domains: either transfer direction maps an RoI from domain A (or B) to B (or A), and then an adversarial network is used to confuse the mapping with RoIs in B (or A). In particular, we use a residual structure for the transformation network, where the identity mapping encourages the feature map to capture shared but not object-specific visual patterns and the residual allows feature transformation.

We demonstrate the effectiveness of the proposed \texttt{Shuffle-Then-Assemble} strategy on two benchmarks: VRD~\cite{lu2016visual} and VG~\cite{krishna2017visual}. We observe consistent improvement of using our pre-trained features against various ablative baselines and other state-of-the-art methods. For example, compared to feature maps without pre-training, we can boost the Recall@100 of supervised, weakly supervised, and zero-shot relationship prediction by absolute 4.74\%, 4.42\%, 4.04\%, respectively on VRD, and 4.41\%, 4.2\%, 5.81\%, respectively on VG. 

\section{Related Work}
\subsubsection{Visual Relationships.}
Modeling the object interactions such as verbs~\cite{gupta2008beyond,chao2015hico}, actions~\cite{gupta2009observing,ramanathan2015learning,yao2010modeling}, and visual phrases~\cite{yatskar2016situation,atzmon2016learning,sadeghi2011recognition,desai2012detecting} has been extensively studied in literature. In particular, our relationship model used in this paper follows the recent progress on modeling generic visual relationships, \ie, the (obj1, rel, obj2) triplets detected in images~\cite{lu2016visual,zhang2017visual}. State-of-the-art relationship models fall into two lines of efforts: 1) message passing between the two object features~\cite{yin2018zoom,li2017vip,xu2017scenegraph}, and 2) exploitation of subject-object statistics such as language priors~\cite{lu2016visual,li2017scene,Zhuang_2017_ICCV} and dataset bias~\cite{zellers2017neural,zhang2017relationship,dai2017detecting}. However, they are still limited in the inherent issue of insufficient training triplets due to combinatorial annotation complexity, leading the resultant relationship model to be brittle to rare or unseen compositions. Though weakly-supervised methods~\cite{zhang2017ppr,peyre2017weakly,wei2018revisiting} can reduce the labeling cost, its performance is still far from practical use compared to supervised models. Unlike previous methods, in this paper, we propose to resolve this challenge in pairwise modeling of relationship, that is, given two regions, we want to improve the predicate classification without additional object information and extra supervision. We believe that the improvement can boost most of the above relationship models by replacing their pairwise modeling counterparts with our method.

\noindent\textbf{Unsupervised Feature Learning.}
By exploiting large-scale unlabeled data, unsupervised feature learning methods~\cite{bengio2013representation} learn more generalizable intermediate data representation for solving some other machine learning tasks. Our motivation for visual relationship feature learning follows the common practice: feature transfer in today's computer vision~\cite{yosinski2014transferable}, which fine-tunes a base network which has been pre-trained on other datasets and tasks. Different from the popular auto-encoder fashion~\cite{zhang2017split,donahue2016adversarial}, our strategy is more similar to the recent works on self-supervised training, where the learning objective is to discover the inherent data compositions such as predicting the context of image patches~\cite{doersch2015unsupervised,noroozi2016unsupervised,pathak2016context,ma2017nips,ma2018cvpr,sun2018cvpr}. In particular, we propose to discover the alignment of RoI pairs and pose this discovery into the task of unsupervised domain transfer using adversarial learning~\cite{zhu2017unpaired,kim2017learning,hoffman2017cycada,yi2017dualgan}. Inspired by them, we use a cycle of transformations to remove the trivial alignment caused by mode collapse and thus build non-trivial connections between the paired RoIs.

\begin{figure}[t!]
\centering
\includegraphics[width=1\linewidth]{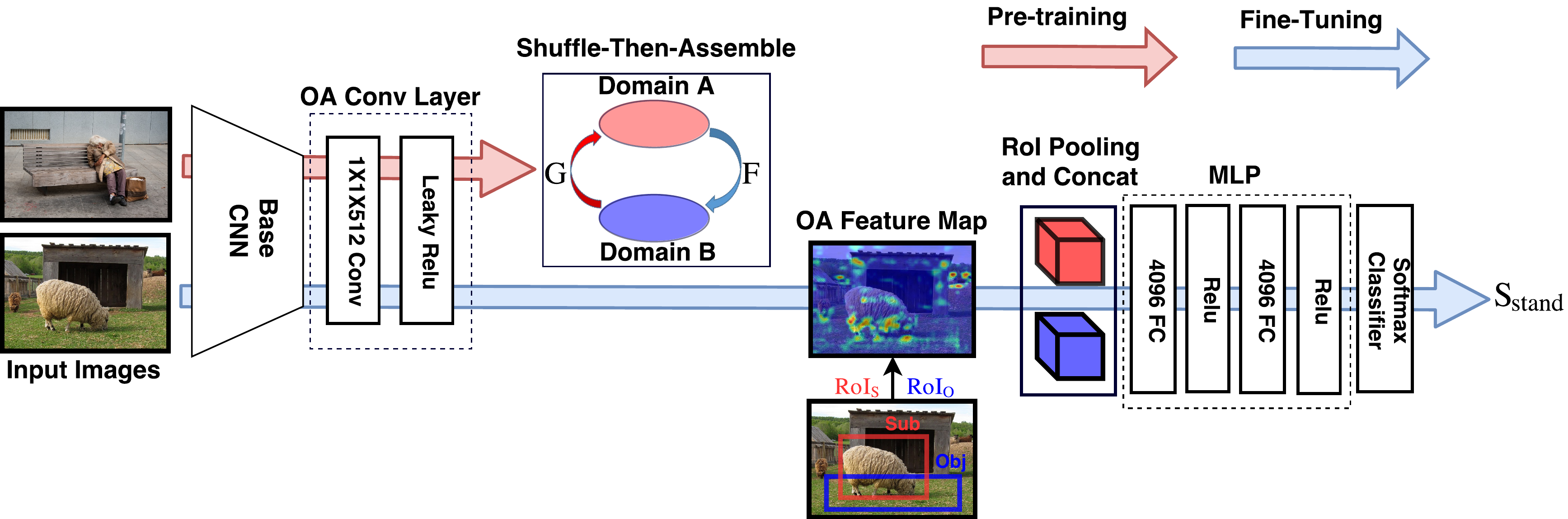}
\caption{The overview of the proposed \texttt{Shuffle-Then-Assemble} pre-training strategy (red arrow) and relationship detection model (blue arrow). The goal at the pre-training stage is to learn the Object-Agnostic (OA) conv-layers using \texttt{Shuffle-Then-Assemble} objective. Then, the traditional supervised training for the relation model can be considered as the fine-tuning stage using the desired OA feature map.
}
\label{fig:2}
\end{figure}
\section{Method}
Fig.~\ref{fig:2} illustrates the overview of using \texttt{Shuffle-Then-Assemble} to enhance the relationship model. The goal of the feature learning process is to pre-train the Object-Agnostic (OA) conv-layers, which result in the desired OA feature map for better relationship modeling. We will first introduce the widely-used relationship modeling framework and its limitations, and then detail how to use the proposed feature learning method to overcome them.

\subsection{Visual Relationship Model}
The input of the visual relationship model is an image with a pair of object bounding boxes, and the output is an ``obj1-rel-obj2'' triplet, where ``obj1'' and ``obj2'' are the object classes of the two bounding boxes, and ``rel'' is the relationship class. In this paper, we adopt the common practice as in~\cite{lu2016visual,zhang2017visual} that we do not directly model the triplet composition as a whole~\cite{sadeghi2011recognition,dai2017detecting}, which requires $\mathcal{O}(C^2R)$ complexity for $C$ object and $R$ relationship classes; instead, we model objects and relationships separately to reduce the complexity down to $\mathcal{O}(C+R)$. Therefore, without loss of generality, we refer to a relationship model as an $R$-way classifier. 

Suppose $\mathbf{x}_i$ and $\mathbf{x}_j$ are the RoI features of any pair of object bounding boxes $(i,j)$ (\eg, the red and blue cubes in Fig.~\ref{fig:2} by RoI pooling~\cite{girshick2015fast}), the $r$-th relationship score is obtained by a softmax classifier whose input is a simple concatenation of the two features:
\begin{equation}\label{eq:1}
S(i,j, r;\theta) = \frac{\textrm{exp}\left(\mathbf{w}_r^T\textrm{MLP}([\mathbf{x}_i,\mathbf{x}_j])\right)}{\sum\limits_{t = 1}^{R}\textrm{exp}\left(\mathbf{w}^T_t\textrm{MLP}([\mathbf{x}_i,\mathbf{x}_j])\right)},
\end{equation}
where $\mathbf{w}_t\in\theta$ is the parameter of the classifier and the configuration of $MLP(\cdot)$ is detailed in Fig.~\ref{fig:2}. Note that although Eq.~\eqref{eq:1} is a naive model and there are fruitful ways of combining $\mathbf{x}_i$ and $\mathbf{x}_j$ in the literature, such as appending independent MLPs for each RoI~\cite{zhang2017visual}, the union RoI~\cite{li2017vip}, and even the fusion with textual features~\cite{hu2016modeling}, our feature learning can be seamlessly incorporated into any of them. We will leave the evaluations of applying these tweaks for future work.

The relationship model can be trained by minimizing the cross-entropy loss of Eq.~\eqref{eq:1}, summing over all the relationship pairs. However, due to the limited annotation of the relationship triplets, relationship models trained on these extremely long-tailed annotations are inevitably biased to the dominant object classes. One may wonder why it is object-biased as Eq.~\eqref{eq:1} does not use any object class information at all? The reason resides in the base CNN feature map. Almost all state-of-the-art visual recognition systems deploy the base CNN~\cite{szegedy2017inception,simonyan2014very,he2016deep} pre-trained on ImageNet~\cite{deng2009imagenet} or ImageNet+MSCOCO~\cite{lin2014microsoft}, where the training task is object recognition. Therefore, the resultant feature map for extracting RoI will naturally favor the sensitivity to object classes --- each RoI feature encodes the discriminative information of the object inside the RoI (cf. the original feature map of Fig~\ref{fig:2}), and leads the parameters in Eq.~\eqref{eq:1} over-fitted to specific object patterns. For example, if most of the triplets of ``stand on'' is ``person stand on street'', then the ``stand on'' classifier will mistakenly consider the joint pattern ``person'' and ``street'' into ``stand on'', and fails in cases of ``person stand on chair'' or ``dog stand on street''.

\begin{figure}[t!]
\centering
\includegraphics[width=1\linewidth]{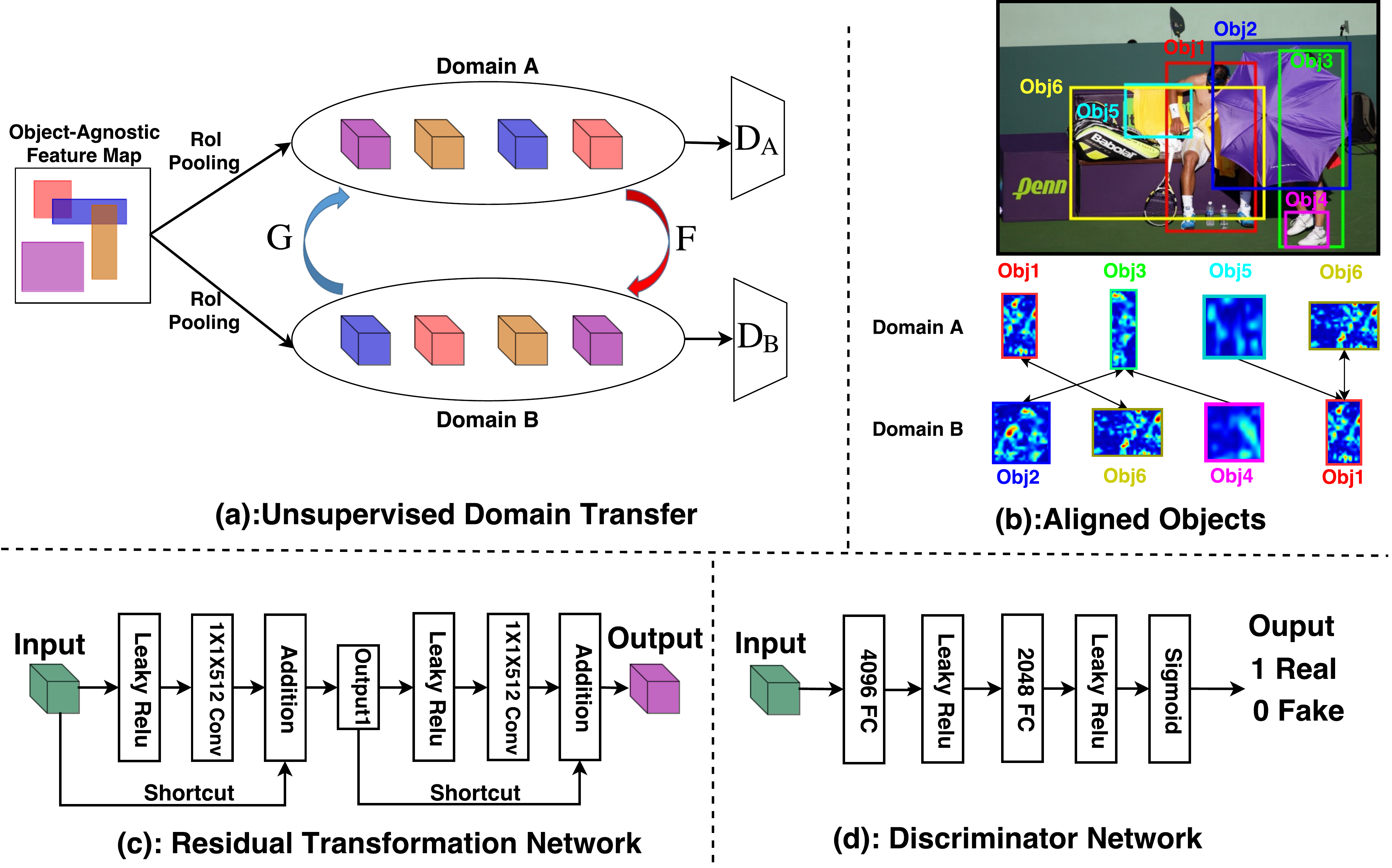}
\caption{(a) The overview of unsupervised domain transfer for \texttt{Shuffle-Then-Assemble}. It contains a cycle of transformations $F$: $A\mapsto B$ and $G$: $B\mapsto A$, and a pair of discriminators $D_A$ and $D_B$ to measure the quality of the transfer. (b) Qualitative transfer results. The directed arrow indicates the nearest-neighbor RoI in the target domain to the RoI from the source domain. (c) The residual architecture of the transformation network. (d) The architecture of the discriminator.
}
\label{fig:3}
\end{figure}

\subsection{Shuffle-Then-Assemble Feature Learning}
To alleviate the bias, we detail our proposed \texttt{Shuffle-then-Assemble} strategy to pre-train the Object-Agnostic (OA) conv-layers for obtaining the OA feature map. As discussed above, the bias is mainly due to the dominant object pairs in training data, therefore, our key idea is to discard the original one-to-one pairwise annotations, \ie, ``shuffle'', leaving two unaligned domains of RoIs for ``obj1'' and ``obj2'', and then we attempt to recover the one-to-one alignment, \ie, ``assemble'', by unsupervised domain transfer. Note that this pre-training strategy does not require additional cost of supervision. As shown in Fig.~\ref{fig:3} (b), we manage to align potential relationships without any one-to-one supervision, \eg, obj1 may relate to obj6 with respect to ``sit'' and obj3 may relate to obj2 with respect to ``hold''.

The unsupervised domain transfer method used in \texttt{Shuffle-Then-Assemble} follows recent progress on adversarial domain transfer~\cite{zhu2017unpaired,kim2017learning,hoffman2017cycada,yi2017dualgan,chen2018zero}. Noteworthy, the motivation of using adversarial domain transfer emphasizes more on the unsupervised alignments but NOT the feature transfer as in traditional domain transfer applications such as~\cite{tsai2018learning}, where the domain transfer is used to close the gap between conveniently available synthetic data and real data. Here, our idea is more similar to~\cite{radford2015unsupervised} which discovers alignments between images that are very visually different such as ``spotted bags'' and ``spotted shoes'', or ``frontal faces'' and ``frontal cars''.

As illustrated in Fig~\ref{fig:3} (a), we want to guide the pre-training of the OA conv-layers by learning mapping functions between domain $A$ and $B$, where each of them consists of RoI features, $\mathbf{a}\in A$ and $\mathbf{b}\in B$, extracted from the tentative OA feature map. For the purpose of domain transfer, we have a cycle of two mappings: $F$: $A\mapsto B$ and $G$: $B\mapsto A$, to discover the underlying relationship between $A$ and $B$. Recall that there is one-to-one supervision between the two domains, we adopt the adversarial objective $\mathcal{L}_{adv}$ such that the mapped features $\{F(\mathbf{a})\}$ and $\{G(\mathbf{b})\}$ are indistinguishable from $B$ and $A$, respectively; in particular, the indistinguishability is measured by two discriminators $D_A$ and $D_B$:
\begin{equation}\label{eq:2}
\begin{split}
&\mathcal{L}_{adv}(A, B; \phi, F, G, D_A, D_B) = \\
&\underbrace{\mathbb{E}_\mathbf{a}[\log D_A(\mathbf{a})] + \mathbb{E}_\mathbf{b}[\log D_B(\mathbf{b})]+\overbrace{\mathbb{E}_\mathbf{b}[\log(1-D_A(G(\mathbf{b}))]+\mathbb{E}_\mathbf{a}[\log(1-D_B(F(\mathbf{a}))]}^{\text{minimize by $F$ and $G$}}}_{\text{maximize by $D_A$ and $D_B$}},
\end{split}
\end{equation}
where $\phi$ is the OA conv-layers that generate $A$ and $B$, $D_A$ is a binary classifier that tries to classify $D_A(\mathbf{a})\mapsto 1$ and $D_A(F(\mathbf{b}))\mapsto 0$, and $D_B$ is defined similarly. In this adversarial way, we will eventually obtain $F$ and $G$ that discover the hidden alignment between the two domains, \ie, indistinguishable by the discriminators.

To encourage more explorations of the potential relationship alignments between the RoIs in the two domains, \eg, avoid from mapping many RoIs in $A$ to only one RoI in $B$ with respect to a trivial spatial relationship such as ``on'' and ``by'', we further impose the ``cycle-consistent'' loss to be minimized by $G$ and $F$:
\begin{equation}\label{eq:3}
\mathcal{L}_{cycle}(A, B; \phi, F, G) = \mathbb{E}_\mathbf{a}[\|\mathbf{a}-G(\mathbf{b})\|_1]+\mathbb{E}_\mathbf{b}[\|\mathbf{b}-F(\mathbf{a})\|_1].
\end{equation}
The loss penalizes two different RoIs, \eg, $\mathbf{a}$ and $\mathbf{a}'$,  mapped to the same RoI $\mathbf{b}$ as it is hard to satisfy $\mathbf{a}\approx G(\mathbf{b})$ and $\mathbf{a}'\approx G(\mathbf{b})$ simultaneously.

By putting Eq.~\eqref{eq:2} and Eq.~\eqref{eq:3} together, the full objective for pre-training the OA conv-layers is:
\begin{equation}\label{eq:4}
\phi^* = \arg\min\limits_\phi\min\limits_{F, G}\max\limits_{D_A, D_B}\mathcal{L}_{adv}(A, B; \phi, F, G, D_A, D_B)+\lambda\mathcal{L}_{cycle}(A, B; \phi, F, G),
\end{equation}
where $\lambda>0$ is a trade-off hyper-parameter. Then, we can use $\phi^*$ to obtain $\mathbf{x}_i$ and $\mathbf{x}_j$, and fine-tune a better relationship model $\theta$ using existing triplet supervision as in Eq.~\eqref{eq:1}. Next, we will introduce the proposed implementation of $F$ and $G$.

\subsection{Implementation Details}
\noindent\textbf{Network Architecture.}
For base CNN, we adopt Faster RCNN (VGG16)~\cite{ren2015faster}, which takes short width to be 600 and outputs the original $1/16\times 1/16 \times 512$ feature map. As shown in Fig~\ref{fig:2}, our OA conv-layer has 1 filter of the size $1\times 1$, stride 1, followed by a Leaky Relu~\cite{xu2015empirical}. The transformation network is detailed in Fig~\ref{fig:3} (c). Each transformation contains two blocks of residual network. The motivation of applying the residual structure is two-fold. 1) The shortcut encourages to find shared regions of two RoIs, since the shared RoI features will pass directly via the shortcut. This makes the optimization not only more light-weighted, but also easier to find the intrinsic inter-related visual patterns as relationships. 2) If any object-specific information is still encoded in the RoI feature, the shortcut will make it harder to achieve the final domain transfer as domain A and B usually contain diverse objects. The discriminator network is detailed in Fig~\ref{fig:3} (d), which is composed by two fully-connected layers followed by Leaky Relu. It takes a 50,176-d (two $ 7 \times 7 \times 512$ RoI feature) vectorized RoI feature as input and outputs a sigmoidal scalar between 0 and 1.

\noindent\textbf{Training.}
At the feature pre-training stage, to collect sufficient RoIs in each domain, we augment the number of original bounding boxes by additional ones with IoU larger than 0.7, extracted by using the Region Proposal Network~\cite{ren2015faster}. For each original bounding boxes, 10 RoIs are sampled. To stabilize the adversarial training in Eq.~\eqref{eq:4}, we adopt three practices:  1) We apply least-square GAN~\cite{mao2017least} to replace the negative log likelihood by a least square loss. 2) The optimizer for training $D_{A}$ and $D_B$ is set to SGD and the optimizer for $G$, $F$ and $\phi$ is set to Adam~\cite{kingma2014adam}. The initial learning rate is set to 1e-4 for both optimizers. 2) $D_{A}$ and $D_B$ are trained three times more compared with $G$, $F$ and $\phi$. The trade-off $\lambda$ in Eq.~\eqref{eq:4} is set to 10. Every mini-batch is one image with randomly selected 128 triplets. The epochs for training these networks are set to 20 on VRD dataset and set to 5 on VG dataset.

At the fine-tune stage for training relationship classifier, the short width of image is still set to 600. Every mini-batch is one image with 128 randomly selected triplets. The optimizer is Adam with initial learning rate set to 1e-5 in all the experiments. The epochs are set to 50 and 30 on VRD dataset and VG dataset, respectively.

\section{Experiments}
We evaluated our \texttt{Shuffle-Then-Assemble} method by performing visual relationship prediction on two benchmark datasets. We conducted experiments under extensive settings: supervised, weakly-supervised, and zero-shot, each of which has various ablative baselines and state-of-the-art methods. We also visualized qualitative object-agnostic features maps compared against others.

\subsection{Datasets and Metrics}
We used two publicly available datasets: VRD (Visual Relationships Dataset\cite{lu2016visual}) and VG (Visual Genome V1.2 dataset~\cite{krishna2017visual}).

\noindent\textbf{VRD dataset.} It contains 5,000 images with 100 object categories and 70 relationships. In total, VRD contains 37,993 relationship triplet annotations with 6,672 unique triplets and 24.25 relationship per object category. We followed the same train/test split as in~\cite{lu2016visual}, \ie, 4,000 training images and 1,000 test images, where 1,877 triplets are only in the test set for zero-shot evaluations.  

\noindent\textbf{VG dataset.} We used the pruned version provided by Zhang\cite{zhang2017visual} since the original one is very noisy. As a result, VG contains 99,658 images with 200 object categories and 100 predicates, 1,174,692 relation annotations with 19,237 unique relations and 57 predicates per object category. We followed the same 73,801/25,857 train/test split. And this dataset contains 2,098 relationships which never occur in the training set, which can be used for zero-shot evaluations. 

\noindent\textbf{Metrics.} As conventions~\cite{lu2016visual,zhang2017visual}, we used Recall@50 (\textbf{R@50}) and Recall@100 (\textbf{R@100}) as evaluation metrics. R@K computes the fraction of times a true relationship is predicted in the top $K$ confident relation predictions in an image. 
\begin{figure}[t!]
\centering
\includegraphics[width=1\linewidth]{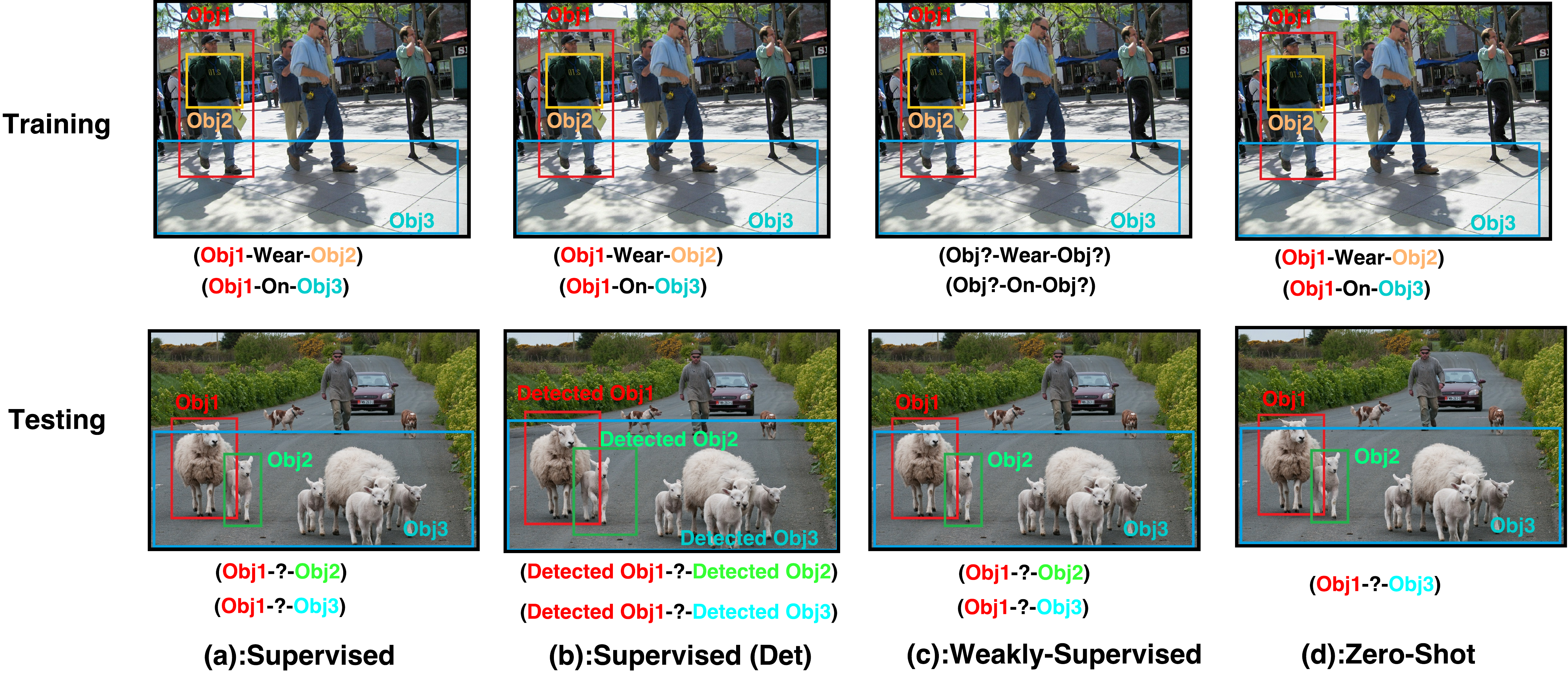}
\caption{ We evaluate relationship prediction task using four different experiment settings: supervised, supervised (Det), weakly-supervised and zero-shot. ``?'' denotes the relationship to be predicted. It is noteworthy that the object category is not know under all the experiment settings, and we only use visual features to predict the relationship between object pairs.
}
\label{fig:4}
\end{figure}

\begin{figure}[t!]
\centering
 \begin{subfigure}{0.8\textwidth} % width of left subfigure
		\includegraphics[width=\textwidth]{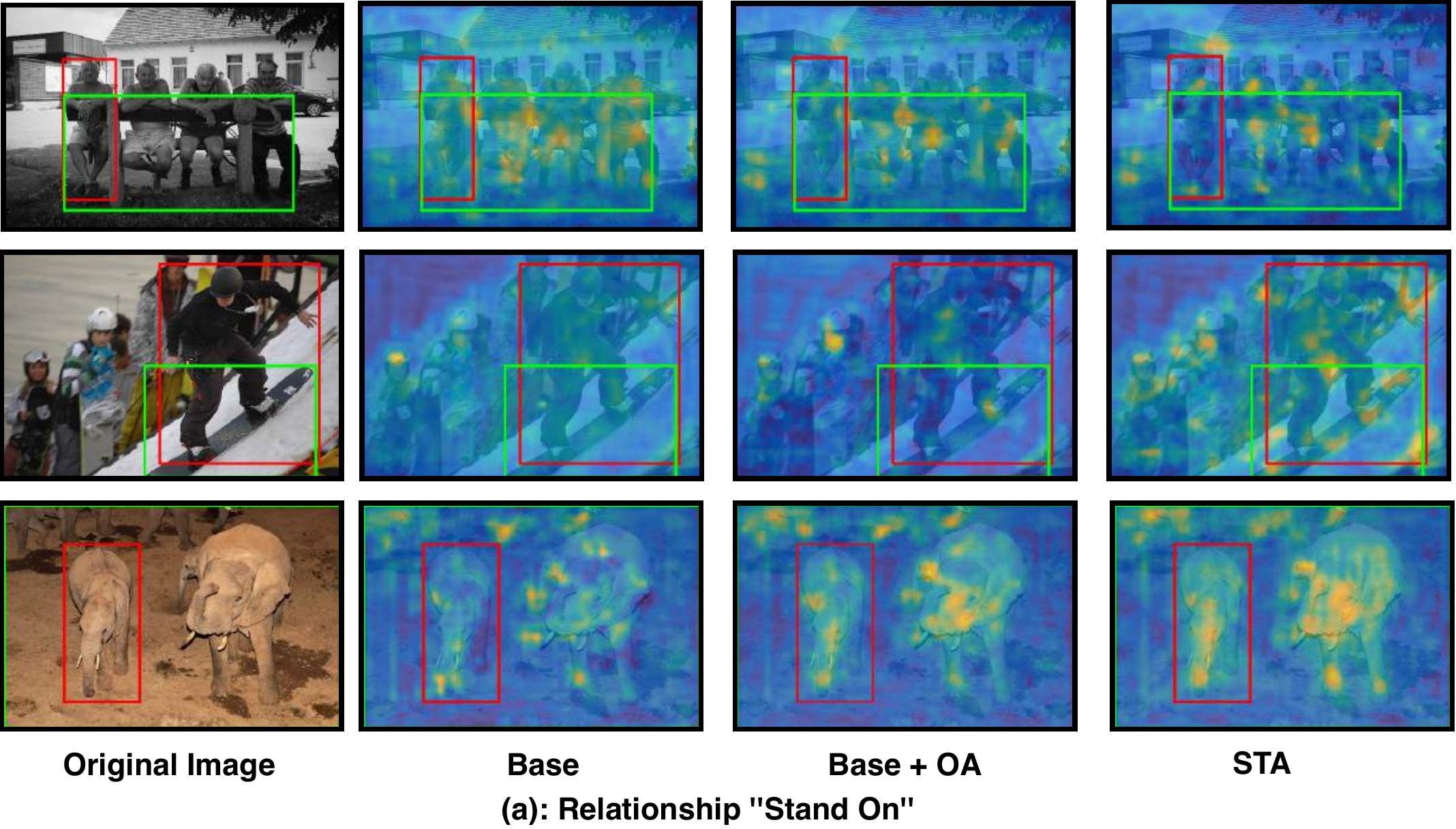}
	\end{subfigure}
	\begin{subfigure}{0.8\textwidth} % width of right subfigure
		\includegraphics[width=\textwidth]{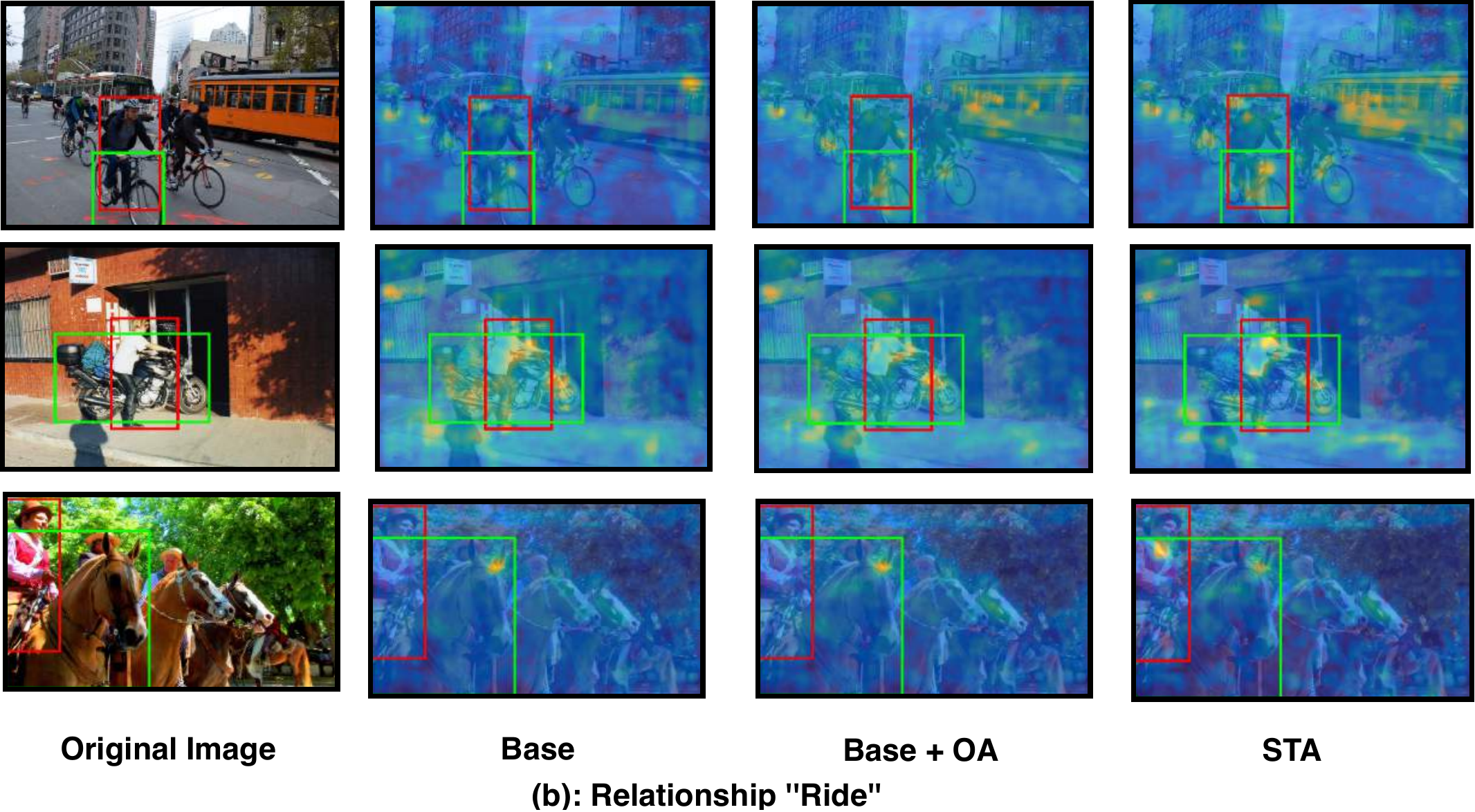}
	\end{subfigure}
\caption{Qualitative feature maps of two relationships on VRD dataset. For each one, three kinds of feature maps are visualized by averaging over the 512 channels. We can see that by using the proposed \texttt{Shuffle-Then-Assemble} (STA), the RoI features are less likely biased to the objects and more focused on the regions of interaction of the two objects. Moreover, the observation is consistent with diverse relationship appearances.
}
\label{fig:5}
\end{figure}

\subsection{Settings}
In our experiments, we only focused on the relationship prediction task, \ie, classifying any two object regions into relationship classes. The reasons are two-fold. First, relationship prediction plays the core role in relationship detection, a more comprehensive task that also needs to detect the two objects. Second, we can exclude the influence of object detection performance, as the improvement of object detection can improve the relationship detection scores~\cite{zhang2017visual}. To offer a testbed for application domains of relationship prediction, we designed the following 4 settings according to different pairwise modeling fashions:

\textbf{Supervised.}
This setting is the standard supervised relationship prediction. As shown in Fig~\ref{fig:4} (a), for training, all the objects are provided with ground truth boxes and the relationship between objects are given; for testing, a pair of objects with bounding boxes are given and their relationship is to be predicted.

\textbf{Supervised (Det)}. 
The above setting assumes a perfect object bounding box detector at testing. However, as shown in Fig~\ref{fig:4} (b), a more practical setting is to use detected object bounding boxes using off-the-shelf object detectors. We used Faster RCNN~\cite{ren2015faster} to detect around 100 objects in an image.

\textbf{Weakly-Supervised}.
Compared to Supervised setting, we discard the one-to-one paired object annotation with respect to a relationship. As shown in Fig~\ref{fig:4} (c), at training, given objects with boxes, we do not know which object relates to which one. Therefore, we used an average-pooling image-level relationship loss:
\begin{equation}
\mathcal{L}_{weak} = -\sum_{i=1}^{N}\sum_{j=1}^N{\sum_{r=1}^R[{ y_{ijr} \log S(i,j,r) + (1-y_{ijr})\log(1-S(i,j,r)) ]}};
\end{equation}
where $N$ is the number of objects,  $y_{ijr}$ is 1 if the object pair $(i,j)$ has the $r$-th relationship, and $S(i,j,r)$ is the relationship score in Eq.~\eqref{eq:1}. Note that the testing stage of this setting is the same as that of Supervised setting. 

\textbf{Zero-Shot}.
This setting is the same as Supervised setting except that at testing we want to predict object pairs whose triplet combination is unseen during training. As shown in Fig~\ref{fig:4} (d), though object sheep, road, and relationship on are individually seen at training, but the composition ``sheep on road'' is novel to test. 

\noindent\textbf{Comparing Methods.}
We compared the proposed \texttt{Shuffle-Then-Assemble} (\textbf{STA}) pre-training strategy with the following ablative baselines: 

\textbf{Base.}
We directly use RoI features which extracted from the base CNN for relationship prediction task.

\textbf{Base+OA.}
We do not pre-train OA conv-layers $\phi$ (in Eq.~\eqref{eq:2}) by \texttt{Shuffle-} \texttt{Then-Assemble} strategy and directly fine-tune $\phi$ and  $MLP(\cdot)$ (in Eq.~\eqref{eq:1}) by minimizing the cross-entropy loss of Eq.~\eqref{eq:1}.

\textbf{STA w/o FT.}
After pre-training $\phi$ by \texttt{Shuffle-Then-Assemble} strategy, the parameters of $\phi$ (in Eq.~\eqref{eq:2}) are fixed. When training the network by minimizing Eq.~\eqref{eq:1}, only parameters of $MLP(\cdot)$ (in Eq.~\eqref{eq:1}) are updated.

\textbf{STA w/o Res.}
The transformation network in Fig~\ref{fig:3} is not a residual network. And the other settings are the same with STA.

We also compared with state-of-the-art visual relationship prediction methods such as \textbf{VTransE}~\cite{zhang2017visual}, \textbf{Lu's-V}~\cite{lu2016visual}, \textbf{Lu's-VLK}~\cite{lu2016visual}, and \textbf{Peyre's-A}~\cite{peyre2017weakly}. Note that except for Lu's-VLK which is a multimodal model, all the methods compared here are visual models. 

\begin{figure}[t!]
\centering
\includegraphics[width=0.8\linewidth]{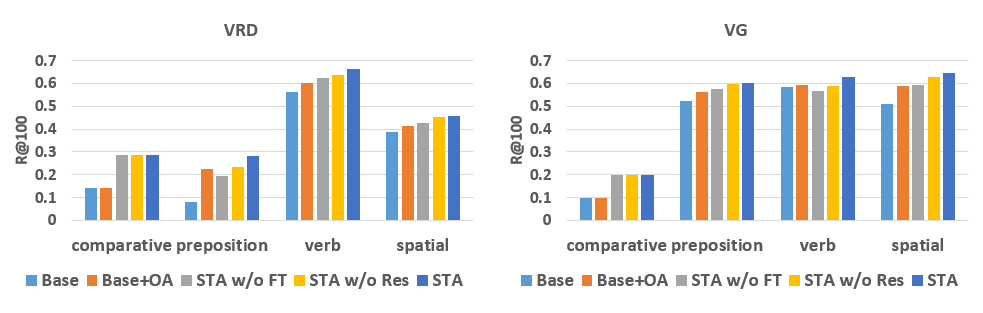}
\caption{ Performances (R@100\%) of relation classification of the four relation types using the different methods in the supervised setting.
}
\label{fig:6}
\end{figure}

\begin{table}[t]
\begin{center}
\caption{The performances (Recall@K\%) of compared methods on two datasets under Supervised setting and Supervised (Det) setting.}
\label{table:1}
\scalebox{.8}{
\begin{tabular}{|c|c|c|c|c||c|c|c|c|}
		\hline
		 Dataset & \multicolumn{2}{|c|}{VRD} & \multicolumn{2}{|c||}{VG} & \multicolumn{2}{|c|}{VRD(Det)} & \multicolumn{2}{|c|}{VG(Det)}\\ \hline
		   Metric&  R@50 & R@100 & R@50 & R@100 &  R@50 & R@100 & R@50 & R@100\\ \hline
           Base               & $39.25$ & $39.25$  & $52.48$ & $52.61$ & $37.83$  & $37.83$ & $50.12$ & $50.31$\\ \hline
           Base+OA                 & $43.29$ & $43.29$  & $58.35$ & $58.53$ & $40.78$  & $40.78$ & $57.03$ & $57.31$      \\ \hline
           STA w/o FT                 & $44.30$ & $44.30$  & $58.14$ & $58.32$ & $41.12$  & $41.12$ & $56.88$ & $57.02$       \\ \hline
           STA w/o Res               & 46.83 & 46.83  & $62.08$ & $62.32$ & $44.85$  & $44.85$ & $61.12$ & $61.30$        \\ \hline
           STA             & \textbf{48.03} & \textbf{48.03}  & \textbf{62.71} & \textbf{62.94} & \textbf{45.65}  & \textbf{45.65} & \textbf{61.27} & \textbf{61.51} \\ \hline \hline
           Lu's-V  \cite{lu2016visual}    & $7.11$ & $7.11$  & $  -  $ & $  -  $ & $  -  $ & $  -  $  & $  -  $ & $  -  $ \\ \hline
           Lu's-VLK \cite{lu2016visual}    & $47.87$ & $47.87$  & $  -  $ & $  -  $ & $  -  $ & $  -  $  & $  -  $ & $  -  $\\ \hline
           VTransE  \cite{zhang2017visual} & $44.76$ & $44.76$  & $62.63$ & $62.87$ & $  -  $ & $  -  $  & $  -  $ & $  -  $\\ \hline
           Peyre's-A\cite{peyre2017weakly} & $46.30$ & $46.30$  & $  -  $ & $  -  $ & $  -  $ & $  -  $  & $  -  $ & $  -  $\\ \hline
\end{tabular}
}
\end{center}
\end{table}

\begin{table}[t]
\begin{center}
\caption{The performances (Recall@K\%) of compared methods on two datasets under Weakly Supervised setting (WS) and Zero-Shot setting(ZS).}
\label{table:2}
\scalebox{.8}{
\begin{tabular}{|c|c|c|c|c||c|c|c|c|}
		\hline
		 Dataset & \multicolumn{2}{|c|}{VRD(WS)} & \multicolumn{2}{|c||}{VG(WS)} & \multicolumn{2}{|c|}{VRD(ZS)} & \multicolumn{2}{|c|}{VG(ZS)}\\ \hline
		  Metric &  R@50 & R@100 & R@50 & R@100 &  R@50 & R@100 & R@50 & R@100\\ \hline
           Base               & $29.36$ & $29.36$ & $45.78$ & $46.01$ & $14.10$ & $14.10$ & $11.04$ & $11.04$     \\ \hline
           Base+OA                 & $31.47$ & $31.47$ & $47.46$ & $47.72$ & $16.53$ & $16.53$ & $13.09$ & $13.09$      \\ \hline
           STA w/o FT                 & $32.84$ & $32.84$ & $47.23$ & $47.39$  & $18.24$ & $18.24$ & $13.72$ & $13.72$      \\ \hline
           STA w/o Res               & $35.10$ & $35.10$ & $50.89$ & $51.13$ & $19.01$ & $19.01$ & $18.03$ & $18.03$    \\ \hline
           STA            & \textbf{35.89} & \textbf{35.89} & \textbf{51.73} & \textbf{51.92} & \textbf{20.57} & \textbf{20.57} & \textbf{18.90} & \textbf{18.90} \\ \hline
           \hline
           Peyre's A\cite{peyre2017weakly} & $34.03$ & $34.03$ & $  -  $ & $  -  $ & $16.10$ & $16.10$ & $  -  $ & $  -  $\\ \hline
\end{tabular}
}
\end{center}
\end{table}

\begin{table}[t]
\begin{center}
\caption{Computed overlap ratios (\%) of two kinds of feature maps}
\label{table:3}
\begin{tabular}{|c|c|c|c|c|c|}
		\hline
		 Dataset & OA  & Base CNN & Dataset & OA  & Base CNN\\ \hline
           VRD   & $\textbf{50.27}$ & $42.45$ & VG    & $\textbf{48.50}$ & $41.32$  \\ \hline
\end{tabular}
\end{center}
\end{table}

\subsection{Results and Analysis}
Table~\ref{table:1},~\ref{table:2} show the performance of compared methods on two datasets of different experiment settings. As we can see, the proposed STA has the best performances compared with the other baselines and state-of-the-art on both datasets. For example, compared to the Base+OA, the proposed STA can boost the Recall@100 of supervised, weakly supervised, and zero-shot relationship prediction by absolute 4.75\%, 4.42\%, 4.04\%, respectively on VRD, and 4.41\%, 4.2\%, 5.81\%, respectively on VG.

Comparing the results of Base+OA with Base, we can see that by adding OA conv-layers, the performance is improved. This observation is basically as expected since the number of parameters have been increased and thus the representation ability of the whole network is improved. By comparing the performance of STA w/o FT with Base+OA, we can find that, even OA conv-layers are not fine-tuned, the features which are pre-trained by \texttt{Shuffle-Then-Assemble} still have comparable performance with the Base+OA. And when the pre-trained OA conv-layers are further fine-tuned (STA w/o Res, STA), the performances will have a considerable boost. Such observations show that the success of the proposed method is not only due to the added small network (OA conv-layers), but also thanks to the proposed \texttt{Shuffle-Then-Assemble} pre-training strategy. 

Fig.~\ref{fig:6} shows the R@100 of relationship prediction of the four relation types which are comparative, preposition, verb and spatial. From this, we can see that the proposed STA has the best performance in each relationship type on both datasets. 

\textbf{Analysis of feature maps.}
Fig.~\ref{fig:5} shows six qualitative examples of feature maps generated by three different strategies. By comparing the STA's feature maps with Base and Base+OA, we can find that STA's feature maps focus more on the overlap regions between subjects and objects. For example, in the second row, STA's feature maps put more attention on people's feet, which would provide cues for predicting the right relationship ``stand on''.

The ratio: $(\sum\nolimits_{i \in R_{over}}{f(i)})/({\sum\nolimits_{i \in R_{joint}}f(i)})$, in Table~\ref{table:3}, is used to measure how our model can focus on the overlapped region. In this formula, $f(\cdot)$ is the normalized joint feature map of subject and object region , $R_{over}$ and $R_{joint}$ mean the overlapped region and the joint region of that feature map respectively. We compare the ratios computed by OA feature and Base CNN feature on both VRD and VG datasets. From the results we can see that the proposed \texttt{Shuffle-Then-Assemble} pre-training strategy can help the relationship model captures more attention on the shared regions between subject and object.

\begin{figure}[t!]
\centering
\includegraphics[width=1\linewidth]{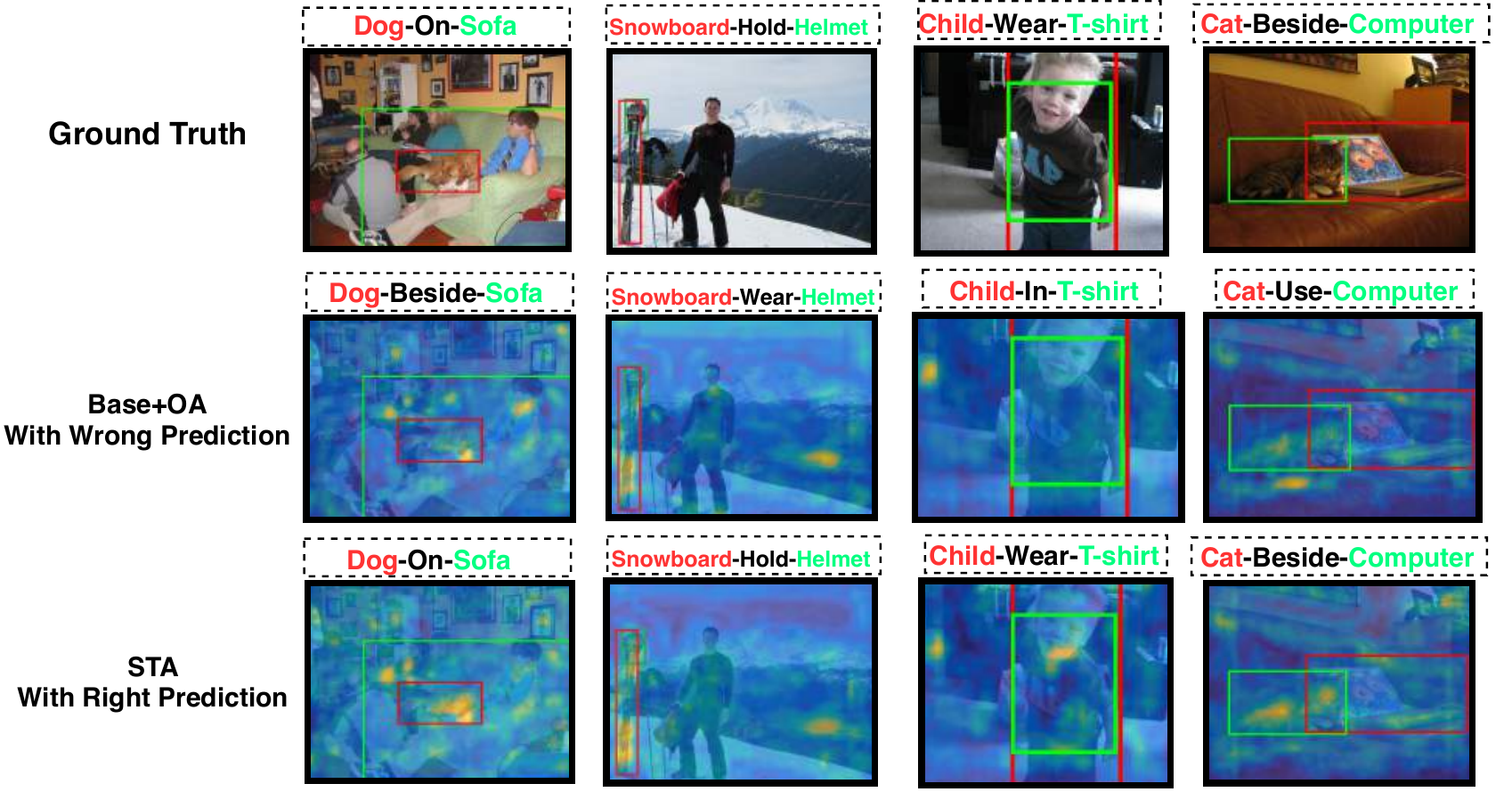}
\caption{Qualitative feature maps of four zero-shot relationships on VRD dataset. For each one, two feature maps of Base+OA with wrong prediction and STA with correct prediction are visualized by averaging over the 512 channels. We can see that by using the proposed \texttt{Shuffle-Then-Assemble} (STA), the RoI features are less likely biased to the objects and more focused on the regions of interaction of the two objects.
}
\label{fig:7}
\end{figure}

\textbf{Analysis of Zero-Shot Setting.}
From table~\ref{table:2}, we can see that the proposed STA has the best performance on both datasets compared with other baselines and one state-of-the-art. This result can further validate the effectiveness of the proposed \texttt{Shuffle-Then-Assemble} pre-training strategy. From the qualitative examples in Fig.~\ref{fig:7}, we can demonstrate that the reason why STA achieves better performance is due to the learned OA feature maps.

\begin{figure}[t!]
\centering
\includegraphics[width=1\linewidth]{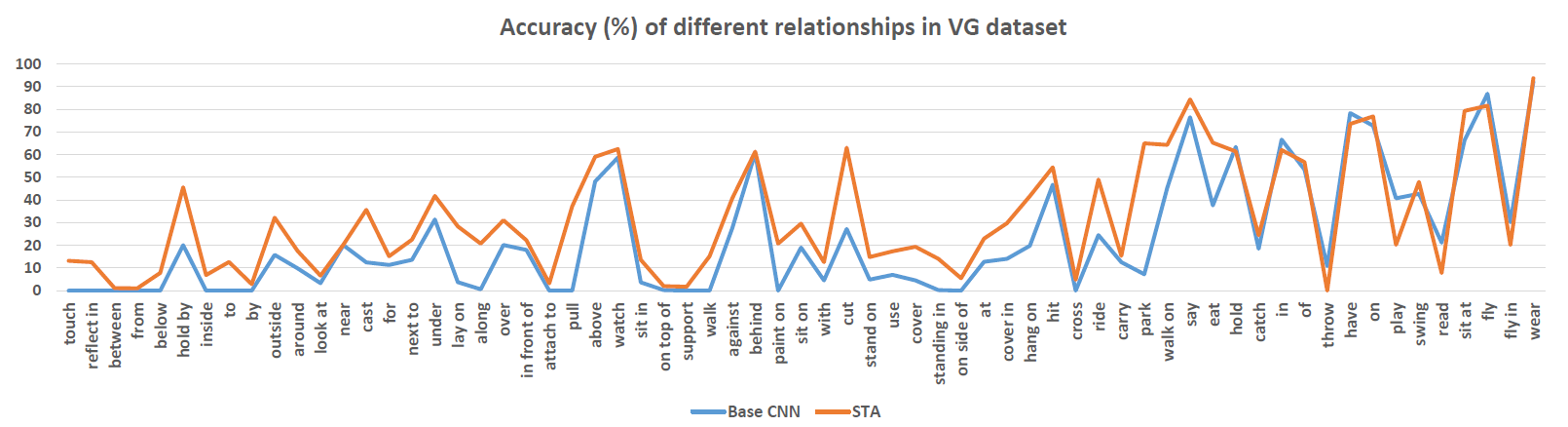}
\caption{The accuracy (\%) of each relationship in VG dataset. In the horizontal axis, the relationships are listed in an ascending order (from left to right) of their biases to specific object combinations. The vertical axis is the accuracy (\%) of each relationship. We can see that for relationships which are less biased to specific combination (left parts), our STA method usually have better performance. 
}
\label{fig:8}
\end{figure}

\textbf{Analysis of object-biased relationships.}
Fig.~\ref{fig:8} shows the accuracy of each relationship, listed in an ascending, left-right order according to their biases to specific subject-object configuration by $N_R(r)/N_C(r)$, where $N_C(r)$ is the number of configurations and $N_R(r)$ is the number of training samples of the $r$-th relationship. Notice that smaller bias indicates more flexible configurations (\textit{e.g.}, ``touch'') and vice versa (\textit{e.g.}, ``wear''). We can find that for relationships which are less biased to specific configurations (left and middle parts), our STA is better as it focuses on object-agnostic features.

\textbf{Failure mode.}
Our model will fail when one relationship depends heavily on specific object combinations. For example, for some relationship listed in the right part of Fig.~\ref{fig:8} (like the relationship ``read'', the subjects and objects are usually ``person'' and ``book''), our model will not defeat the baseline. Under this condition, the object categories will be useful for predicting relationship. Note that such failure can be easily recovered by rules mined from dataset statistics.

\section{Conclusions}
We proposed a novel \texttt{Shuffle-Then-Assemble} visual relationship feature learning strategy for improving visual relationship models. The key idea is to discard the original one-to-one paired object alignments, and then try to recover them in an unsupervised pair discovery fashion by using a cycle-consistent adversarial domain transfer method. In this way, the object class information in object pairs is excluded and hence the resultant feature map is less likely biased to specific object compositions. On two visual relationship benchmarks, we found a consistent improvement from a naive relationship prediction model using the pre-trained OA feature maps. 

\noindent\textbf{Acknowledgments.}
This research is partially support by NTU-CoE Grant, Alibaba-NTU JRI, and Data Science \& Artificial Intelligence Research Centre@NTU (DSAIR).
\clearpage

\bibliographystyle{splncs04}
\bibliography{1372}
\end{document}